\documentclass[a4paper]{article}
\usepackage[ruled,vlined]{algorithm2e}
\usepackage{graphicx}
\usepackage{INTERSPEECH2019}
\usepackage{multirow}
\usepackage{colortbl}
\usepackage{diagbox}

\title{Completely Unsupervised Speech Recognition By A Generative Adversarial Network Harmonized With Iteratively Refined Hidden Markov Models}
\name{Kuan-Yu Chen, Che-Ping Tsai, Da-Rong Liu, Hung-Yi Lee, Lin-shan Lee}
\address{
  Graduate Institute of Communication Engineering, National Taiwan University}
\email{\{r06942070, r06922039, r07942148\}@ntu.edu.tw, tlkagkb93901106@gmail.com, lslee@gate.sinica.edu.tw}

\begin{document}

\maketitle
\begin{abstract} 
Producing a large annotated speech corpus for training ASR systems remains difficult for more than 95\% of languages all over the world which are low-resourced, but collecting a relatively big unlabeled data set for such languages is more achievable. This is why some initial effort have been reported on completely unsupervised speech recognition learned from unlabeled data only, although with relatively high error rates. In this paper, we develop a Generative Adversarial Network (GAN) to achieve this purpose, in which a Generator and a Discriminator learn from each other iteratively to improve the performance. We further use a set of Hidden Markov Models (HMMs) iteratively refined from the machine generated labels to work in harmony with the GAN. The initial experiments on TIMIT data set achieve an phone error rate of 33.1\%, which is 8.5\% lower than the previous state-of-the-art.

\end{abstract}


\section{Introduction}
Automatic speech recognition (ASR) has achieved remarkable performance and been widely used. 
However, the state-of-the-art ASR systems \cite{chorowski2015attention,chiu2018state} have to learn from massive annotated data which is difficult to obtain for at least 95\% of the languages over the world which are low-resourced. Conversely, collecting a relatively big unlabeled corpora for such languages is more achievable in the big data era. This is why unsupervised ASR is attractive.

Substantial effort was made to learn signal representations directly from speech signals in an unsupervised way, which could be a good step towards unsupervised ASR, and has been shown useful in various tasks,  such as speaker identification \cite{dehak2009support}, spoken term detection \cite{lee2013enhanced, chen2013hybrid, norouzian2012exploiting}, and spoken document retrieval \cite{levin2013fixed, levin2015segmental, kamper2016deep}. 
In particular, a sequence-to-sequence auto-encoder \cite{cho-al-emnlp14,sutskever2014sequence} was used to embed audio segments into vectors of fixed-dimensionality. 
Inspired by Word2Vec \cite{mikolov2013efficient}, it was shown possible to encode some semantics into such audio embeddings \cite{chung2018speech2vec}. But there still exists a wide gap between all these works and unsupervised ASR.

On the other hand, unsupervised neural machine translation was very successful recently \cite{artetxe2018unsupervised, conneau2018word, lample2018unsupervised}, in which a mapping relationship between source and target language word embedding spaces could be learned with adversarial training \cite{goodfellow2014generative} in an unsupervised manner. This led to several attempts on unsupervised ASR, since ASR is also a kind of translation. This included aligning audio and text embedding spaces for the purpose of unsupervised ASR \cite{chung2018unsupervised, chen2018phonetic,chen2018almost}, and our prior work \cite{liu2018completely} of unsupervised phoneme recognition, with a Generative Adversarial Network (GAN) \cite{goodfellow2014generative}, by clustering audio embeddings into a set of tokens, and learning the mapping relationship between tokens and phonemes.

In the above efforts, it was realized that the primary difficulties for applying unsupervised neural machine translation model for ASR purposes were the segmental structure of the audio signals, i.e. each word or phoneme consists of a segment of consecutive frames of variable length with unknown boundaries in the signal, and ASR is supposed to map such a segmental structure to a sequence of discrete words or phonemes. This is why oracle or forced alignment audio segmentation boundaries was usually helpful to achieve satisfactory performance in these works.

This above problem was properly handled previously by a specially designed cost function called Segmental Empirical Output Distribution Matching, which considered both the n-gram probabilities across the output units and the intra-segment frame-wise smoothness \cite{yeh2018unsupervised}. However, this approach required a very large batch size during training to avoid being biased \cite{liu2017unsupervised}, and became difficult when the training data size grew. Furthermore, the n-gram probabilities considered here only included local statistics in the output sequences, while other information such as long-distance dependency were inevitably ignored.

In this paper, we propose to handle the above problem by a framework using Generative Adversarial Network (GAN) harmonized with a set of iteratively refined hidden Markov models (HMMs). Only unlabeled utterances and unrelated text sentences are needed, but not segment boundaries at all.
The overall framework is shown in Fig. \ref{fig:Architecture}. The GAN includes a generator and a discriminator iteratively learning from each other. The generator consists of two parts: (a) a frame-wise phoneme classifier and (b) a sampling process.
(a) transforms a sequence of acoustic features into a sequence of frame-wise phoneme prediction. Then based on the present segmentation, (b) samples a phoneme prediction from each segment to generate the predicted phoneme sequence.
The discriminator is trained to distinguish the predicted phoneme sequences from the real phoneme sequences obtained from text sentences. 
On the other hand, we use the generator output to train a set of HMM models (not shown in Fig. \ref{fig:Architecture}), and use these HMMs to refine the segmentation on the training set, which are used to learn better generator and discriminator.
The harmonized process of training GAN and HMM models improves the performance iteratively. 

This framework works easily with a very large data set, and the discriminator considers all possible information from the text data set, not limited to the n-gram local statistics. This framework achieved 33.1\% phone error rate (PER) on TIMIT in the preliminary experiments, which is 8.5\% lower than the previous state-of-the-art \cite{yeh2018unsupervised}. 

\section{Proposed framework}
Below we describe the GAN architecture and training loss function in section \ref{model} and section \ref{loss} and the harmonized HMMs in section \ref{refine}.

\subsection{GAN model architecture} \label{model}
\begin{figure}[t]
  \centering
  \includegraphics[width=\linewidth]{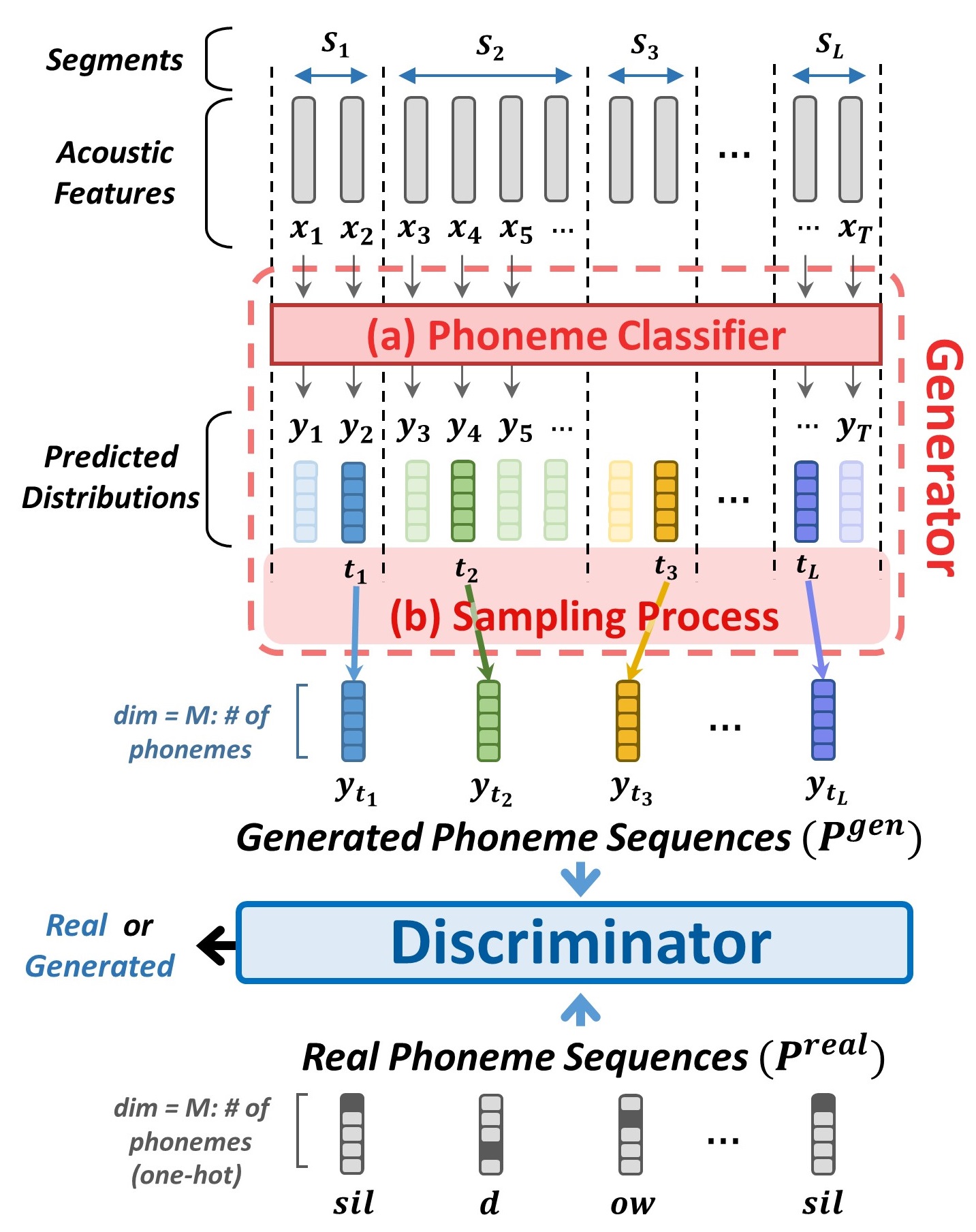}
  \caption{Overview of the proposed approach. The generator includes (a) phoneme classifier transforming  the acoustic features into predicted phoneme distributions, and (b) a phoneme distribution sampled from each segment. The discriminator is trained to distinguish between the generated and real phoneme sequences. The HMMs are not shown.}
  \label{fig:Architecture}
\end{figure}

The Generative Adversarial Networks (GAN) \cite{goodfellow2014generative, arjovsky2017wasserstein, yu2017seqgan} consists of a generator $\mathcal{G}$ and a discriminator $\mathcal{D}$. The discriminator learns to distinguish the generator output from real phoneme sequences, while the generator learns to produce phoneme sequences which can "fool" the discriminator. So the generator and the discriminator learn from each other iteratively. As shown at the middle of Figure \ref{fig:Architecture}, the generator has two parts: a frame-wise phoneme classifier and a sampling process. 

An input feature sequence $\textbf{x}=\{x_1,...,x_{T}\}$ is fed to the phoneme classifier, producing predicted phoneme distribution sequence $\textbf{y}=\{y_1,...,y_{T}\}$, where $x_{t}$ is a $d$-dimensional acoustic feature at time $t$, $y_{t}$ is a probability distribution over all possible $M$ phonemes at time $t$, and $T$ is the input sequence length.
This classifier is a context-dependent DNN network \cite{dahl2012context},  or an Recurrent neural network (RNN).

Sampling process is then applied to address the segmental structure issue. 
Any unsupervised audio segmentation approach can be used to produce an initial segmentation of the input sequence $\textbf{x}$ mentioned above,  $\textbf{S}=\{S_1,...,S_{L}\}$ , where $S_l$ is the $l$th segment and $L$ is the total number of segments. 
We randomly sample a phoneme distribution $y_{t}$ 
from each segment to generate a phoneme distribution sequence, which is referred to as the generated phoneme sequence and denoted as $\uppercase{p}^{gen}=\{y_{t_1},...,y_{t_L}\}$, where $y_{t_l}$ is the phoneme distribution sample from the segment $S_l$. 

The discriminator is to discriminate between the real phoneme sequence $\uppercase{p}^{real}$ (one-hot) from a text data set and the generated phoneme sequence $\uppercase{p}^{gen}$ from the generator giving an output scalar. The higher the scalar, the more probable it is a real phoneme sequence. This discriminator is a two-layer CNN. 

\subsection{Training loss} \label{loss}
There are two objectives for the training loss. First, the generated phoneme distribution sequences should be very close to those of the real phoneme sequences obtained with text data, which is the target of the GAN. Second, the phoneme distributions for frames in the same segment should be very close to one another, which leads to a loss term referred to as intra-segment loss here.

\subsubsection{Discriminator loss}
The loss for training the discriminator $\mathcal{D}$ follows the concept of Wasserstein GAN \cite{arjovsky2017wasserstein} with gradient penalty \cite{NIPS2017_7159}:
\begin{equation}
      \mathcal{L}_{D} = 
      \frac{1}{K} \sum_{k=1}^{K}D(\uppercase{p}^{gen(k)})   
       - \frac{1}{K} \sum_{k=1}^{K}D(\uppercase{p}^{real(k)}) + \alpha \mathcal{L}_{gp},
  \label{discriminator_loss} 
\end{equation}
where $D(P)$ is the scalar output of the discriminator for an input sequence $P$, larger for a real phoneme sequence, $K$ is the number of training examples in a batch, and $k$ is the example index. $\alpha$ is a weight for the gradient penalty $\mathcal{L}_{gp}$.
\begin{equation}
      \mathcal{L}_{gp} = 
      \frac{1}{K} \sum_{k=1}^{K}( ||\nabla D( \uppercase{p}^{inter(k)})||-1 )^2,
  \label{gp_loss} 
\end{equation}
where $\uppercase{p}^{inter}$ is from real phoneme sequence $\uppercase{p}^{real}$ and a generated phoneme sequence $\uppercase{p}^{gen}$ with random weights between 0 and 1, and this term is useful in stabilizing the training.

\subsubsection{Generator loss}
\label{sec:generator_loss}
Different from the original Wasserstein GAN, here we introduced an intra-segment loss:
\begin{equation}
      \mathcal{L}_{intra} = \frac{1}{K} \sum_{k=1}^{K} \sum_{i,j\in \textbf{S}_{k}} 
      (y_{i} - y_{j})^2,
  \label{intra_segment_loss} 
\end{equation}
so the phoneme distributions for frames within the same segment can be more homogeneous. The combined generator loss is:
\begin{equation}
      \mathcal{L}_{G} = -\frac{1}{K} \sum_{k=1}^{K}D(\uppercase{p}^{gen(k)})  + \lambda \mathcal{L}_{intra},
  \label{generator_loss} 
\end{equation}
where $\lambda$ is a weight. 
The generator and the discriminator are iteratively trained to learn from each other, so the phoneme classifier in the generator is eventually able to map acoustic feature sequences to phoneme sequences "looking real".

The inference is then performed when the GAN is well trained. We simply map the acoustic feature sequence $\textbf{x}$ of the training set to  the corresponding phoneme distribution sequences $\textbf{y}$ , pick up the phoneme with the highest probability for each frame, and select the phoneme picked from frames within a segment with the highest probability as the phoneme recognition result. The phoneme recognition result can also be obtained using available decoders such as WFST, which  includes lexicon and language model information. For example, the segmentation boundaries are even not needed in WFST decoders.

\subsection{Harmonization with iteratively refined HMMs} \label{refine}
When the training set is decoded into phoneme sequences by a set of well trained GAN as above, these GAN-generated phoneme sequences are taken as labels for the training set to train a set of phoneme HMMs. This set of phoneme HMMs are then used to re-transcribe the training set by force alignment into new phoneme sequences with new segmentation boundaries, which are then used to start a new iteration of GAN training as described in sections \ref{model} and \ref{loss}, then train a refined set of HMMs as mentioned above. This GAN/HMM harmonization procedure can be performed iteratively, as depicted in Algorithm \ref{alg:algorithm} until converged.

\begin{algorithm}
  \caption{GAN/HMM Harmonization}
  \label{alg:algorithm}
  \KwIn{Real phoneme sequences $\uppercase{p}^{real}$, Speech utterances, initial phoneme segmentation boundaries $b$}
  \KwOut{Unsupervised ASR system}
  \While{not converged}{
    Given $b$, in an unsupervised way train the GAN\;
    Obtain transcriptions $T$ of speech utterances using the generator within the GAN\;
    Given $T$, train the HMMs\;
    Obtain a new $b$ by forced alignment with the HMMs.\
  } 
\end{algorithm}

\section{Experimental Setup}

\subsection{Dataset}
\label{sec:dataset}
The TIMIT corpus \cite{garofolo1993darpa} was used in the preliminary experiments, which included recordings of phonetically-balanced read speech, with 6300 utterances from 630 speakers, with  4620 utterances for training and 1680 for testing.
Each utterance included manually aligned phonetic/word transcriptions, as well as a 16-bit, 16kHz waveform file. 39-dim MFCCs were extracted with utterance-wise cepstral mean and variance normalization (CMVN) applied. 
We selected 4000 utterances in original training set for training and others for validation. 
We further randomly removed 4\% of the phonemes and duplicated 11\% of the phonemes in the real phoneme sequences for the training set to generate augmented phoneme sequences to be used with real phoneme sequences in training the GAN. This is referred to as data augmentation below.

\subsection{Experimental setting}
All models were trained with stochastic gradient descent using a mini-batch size of 150 and Adam optimization. For the intra-segment loss in equation (3),  
we sampled 6 sets of $(y_i,y_j)$ in each segment from each training example.
The phoneme classifier in Fig.~\ref{fig:Architecture} was an one-layer DNN with 512 ReLU units with 48 output classes. 
The input feature was a concatenation of 11 windowed frames of  MFCCs          .
The discriminator was a two-layer 1D CNN, first layer with 4 different kernel sizes: 3, 5, 7 and 9, each with 256 channels, while second layer  with kernel size 3 and 1024 channels.  The weights $(\lambda, \alpha)$  were $(0.5, 10)$.
The learning rate for $\mathcal{G}$ and $\mathcal{D}$ were set to 0.001 and 0.002 respectively. 
Every GAN training iteration consisted of 3 $\mathcal{D}$  updates and a single $\mathcal{G}$ update. 

WFST including an 9-gram phoneme language model was used as mentioned in Section \ref{loss}, where a state represents a phoneme. For modeling state transition probability in the unsupervised setting, we set the self-loop probability to 0.95 and 0.05 to other phonemes for all the phoneme states. HMM (monophone and triphone) training followed the standard recipes of Kaldi \cite{povey2011kaldi}. Linear Discriminant Analysis(LDA) and Maximum Likelihood Linear Transform(MLLT) were applied to MFCCs for model training. The ratio of the acoustic to the language models were set to 1:20 and 1:1 for the phoneme classifier and HMM models, respectively.  The evaluation metrics were phone error rate (PER) and frame error rate (FER) for 39 phoneme classes mapped from the 48 output classes of the classifier. 

\section{Experimental Results}

The first set of results are listed in Table \ref{tab:performance}. The upper section (I) of the table is for the case that all the 4000 training  utterances are well labeled. The middle section (II) is the case that the oracle boundaries provided by TIMIT were used but nothing else. The lower section (III) is the case that the initial boundaries were obtained automatically with GAS \cite{wang2017gate}. The phoneme sequences of the training set were used in two different ways: In the left column of the table labeled "Matched", the phoneme transcriptions in all the 4000 training utterances were used as the real phoneme sequences in GAN training, which means the utterances and the real phoneme sequences are matched but not aligned during training. In the right column labeled "Nonmatched", 3000 utterances were taken as acoustic features while the phoneme transcriptions of the other 1000 utterances taken as the real phoneme sequences, no overlap between the two. All HMMs in this table had the same setting, triphone models with LDA+MLLT features.

\begin{table}[t]
  \caption{Comparison of different methods}
  \label{tab:performance}
  \centering
  \resizebox{\columnwidth}{!}{
  \begin{tabular}{|>{\columncolor[rgb]{0.9,0.9,0.9}}l|>{\columncolor[rgb]{0.9,0.9,0.9}}l|l|c|c|c|c|}
    \hline
    \multicolumn{3}{|c|}{\multirow{3}{*}{Approaches}} & \multicolumn{2}{|>{\columncolor[rgb]{0.9,0.9,0.9}}c|}{Matched} & \multicolumn{2}{|>{\columncolor[rgb]{0.9,0.9,0.9}}c|}{Nonmatched} \\ 
    \multicolumn{3}{|c|}{} & \multicolumn{2}{|>{\columncolor[rgb]{0.9,0.9,0.9}}c|}{(all 4000)} & \multicolumn{2}{|>{\columncolor[rgb]{0.9,0.9,0.9}}c|}{(3000/1000)} \\ \cline{4-7}
    \multicolumn{3}{|c|}{} & FER & PER & FER & PER \\ \hline
    \multicolumn{7}{|>{\columncolor[rgb]{0.9,0.9,0.9}}c|}{(\uppercase\expandafter{\romannumeral1})  Supervised (labeled)} \\ \hline
    \multicolumn{3}{|l|}{(a) RNN Transducer \cite{yeh2018unsupervised}}       &  -   & 17.7 & - & - \\ \hline
    \multicolumn{3}{|l|}{(b) standard HMMs}        &  -   & 21.5 & - & - \\ \hline
    \multicolumn{3}{|l|}{(c) Phoneme classifier}   & 27.0 & 28.9 & - & - \\ \hline
    \multicolumn{7}{|>{\columncolor[rgb]{0.9,0.9,0.9}}c|}{(\uppercase\expandafter{\romannumeral2})  Unsupervised (with oracle boundaries)} \\ \hline
    \multicolumn{3}{|l|}{(d) Mapping relationship GAN \cite{liu2018completely}}   & 40.5 & 40.2 & 43.6 & 43.4    \\ \hline
    \multicolumn{3}{|l|}{(e) Segmental empirical-ODM \cite{yeh2018unsupervised}}    & 33.3 & 32.5 & 40.0 & 40.1    \\ \hline
    \multicolumn{3}{|l|}{(f) Proposed: GAN}              & 27.6 & 28.5 & 32.7 & 34.3    \\ \hline
    \multicolumn{7}{|>{\columncolor[rgb]{0.9,0.9,0.9}}c|}{(\uppercase\expandafter{\romannumeral3})  Completely unsupervised (no label at all)} \\ \hline
    \multicolumn{3}{|l|}{(g) Segmental empirical-ODM \cite{yeh2018unsupervised}}        &   -  & 36.5 &   -  & 41.6    \\ \hline
     &  & (h) GAN & 48.3 & 48.6 & 50.3 & 50.0    \\ \cline{3-7}
     & \multirow{-2}{*}{iteration 1} & (i) GAN/HMM &   -  & 30.7 &   -  & 39.5    \\ \cline{2-7}
     &  & (j) GAN & 41.0 & 41.0 & 44.3 & 44.3    \\ \cline{3-7}
     & \multirow{-2}{*}{iteration 2} & (k) GAN/HMM &   -  & 27.0 &   -  & 35.5    \\ \cline{2-7}
     &  & (l) GAN & 39.7 & 38.4 & 45.0 & 44.2    \\ \cline{3-7}
     \multirow{-6}{*}{\rotatebox{90}{Proposed}} & \multirow{-2}{*}{iteration 3} & (m) GAN/HMM &   -  & 26.1 &   -  & 33.1    \\ \hline
  \end{tabular}
  }
\end{table}
\subsection{Supervised baselines}
In section (I) for supervised approaches with completely unlabeled data, in row (a) the RNN Transducer \cite{graves2012sequence} was very powerful, while the standard triphone HMMs in row (b) were very strong too. The phoneme classifier in row (c) was exactly the phoneme classifier in the generator of Fig.\ref{fig:Architecture}, except trained with annotated transcriptions.

\subsection{Unsupervised but with oracle boundaries}
With the oracle phoneme boundaries provided by TIMIT, rows (d) (e) in the middle section (II) are for two previously reported baselines, while row (f) for exactly the proposed GAN in Fig.\ref{fig:Architecture} except without the harmonized HMMs and data augmentation process mentioned in section \ref{sec:dataset} achieved significantly lower PER and FER. Interestingly, the PER in matched case achieved (28.5\%) in row (f) was even better than the supervised phoneme classifier (28.9\%) in row (c) trained with labeled data, indicating the power of GAN. 
The discriminator of GAN in row (f) considered the whole generated phoneme sequences, while the DNN in row (c) considered only the input acoustic features. We also see the PER gap between matched and nonmatched cases for the proposed GAN in row (f) (5.8\%) is smaller than the prior work of Segmental Empirical-ODM in row (e) (7.6\%). The PER (34.3\%) in nonmatched case by the proposed GAN was also close to the matched case of the prior work of Segmental Empirical-ODM in row (e) (32.5\%).

\subsection{Completely unsupervised} 
In the lowest section (III) of Table \ref{tab:performance}, only the first row (g) is for the prior work of \cite{yeh2018unsupervised}, while all other rows (h)-(m) are for the approaches proposed here, respectively with 1, 2, and 3 iterations of harmonized GAN/HMM, rows (h) (j) (l) for GAN alone not further harmonized with HMMs, while rows (i) (k) (m) further harmonized with HMMs. All these approaches were based on the initial segmentation boundaries automatically generated by GAS \cite{wang2017gate}. 

We see the performance was consistently improved significantly after each iteration for either GAN alone (rows (h) (j) (l)) or GAN/HMM (rows (i) (k) (m)), or after harmonization with HMMs at each iteration (rows (i) v.s. (h), (k) v.s. (j), (m) v.s. (l)), for both matched and nonmatched cases, for both PER and FER. Not to mention the prior work of Segmental Empirical-ODM in row (g) needed a large batch-size (up to 20000 training examples in a batch) to achieve a satisfactory performance, while the training process here was done with a batch size as small as 150.
 
Note that the PER for GAN/HMM after iteration 2 (row (k)) in the matched case (27.0\%) was even lower than all results with oracle boundaries (rows (d) (e) (f)) and supervised phoneme classifier with labeled data (row (c)). The GAN/HMM harmonization algorithm converged after three iterations in the preliminary experiments, ended up with PER of 26.1\% and 33.1\% in matched and nonmatched cases respectively, which were in fact 10.4\% and 8.5\% lower than the prior work in row (g), although still far behind the strong supervised baselines in rows (a) (b). All these verified the power of the proposed harmonized GAN/HMM approach.

\subsection{Ablation studies}
\begin{table}[t]
  \caption{Ablation studies for completely unsupervised model after 1 iteration of GAN/HMM harmonization (row (i) of Table \ref{tab:performance}) in Nonmatched case.}
  \label{tab:ablation}
  \centering
  \resizebox{\columnwidth}{!}{
  \begin{tabular}{|c|l|c|c|}
    \hline
    \multicolumn{2}{|c|}{Completely unsupervised} & \textbf{} & \textbf{}       \\
    \multicolumn{2}{|c|}{(iteration 1, Nonmatched)} & FER & PER \\
    \hline
    \multirow{2}{*}{(1)} & 
    GAN/HMM (tri+LDA+MLLT)   & 
    \multirow{2}{*}{-}   & 
    \multirow{2}{*}{39.5} \\ 
    
     & (row (i) of Table 1) & & \\  \hline
    (2) & GAN/HMM (tri)                                  &  -   & 41.9 \\ \cline{1-4} 
    (3) & GAN/HMM (mono)                                 &  -   & 43.8 \\ \cline{1-4} 
    (4) & GAN (row (h) of Table 1)                       & 50.3 & 50.0 \\ \cline{1-4} 
    (5) & GAN - Augm                                     & 53.6 & 51.9 \\ \cline{1-4} 
    (6) & GAN - Augm - $\mathcal{L}_{intra}$             & 63.0 & 62.6 \\ \cline{1-4} 
    (7) & GAN in (4) with RNN                            & 75.5 & 71.6 \\
    \cline{1-4} 
  \end{tabular}
  }
\end{table}
Ablation studies for the completely unsupervised harmonized GAN/HMM model obtained after iteration 1 in row (i) of Table \ref{tab:performance} in the nonmatched case, initiated with the segmentation boundaries obtained by GAS, are reported in Table \ref{tab:ablation}.

Row (1) in Table \ref{tab:ablation} corresponds to row (i) in Table \ref{tab:performance}, GAN/HMM at iteration 1, in which the HMMs were triphones with LDA and MLLT. Exactly the same except without LDA and MLLT is in row (2), and triphones further replaced by mono-phones in row (3). 
GAN alone without HMM harmonization (exactly row (h) of Table \ref{tab:performance}) is in row (4), from which the data augmentation process mentioned in section \ref{sec:dataset} was removed is in row (5), while the loss $\mathcal{L}_{intra}$ in section \ref{sec:generator_loss} is further removed in row (6). 
Here we see the performance degraded step by step. For the HMMs used here, triphones are obviously better than mono-phones, and LDA and MLLT helped. For the GAN both data augmentation and the intra-segment loss made contributions. When we replaced the DNN used in the phoneme classifier in the GAN in row (4) here or row (h) in Table \ref{tab:performance} by a RNN, the results is in row (7), which showed RNN did not work here, probably because the long term dependency  captured by RNN was able to "fool" the discriminator while generating output unrelated to the input.

\subsection{Comparison with supervised approach}
\begin{figure}[t]
  \centering
  \includegraphics[width=\linewidth]{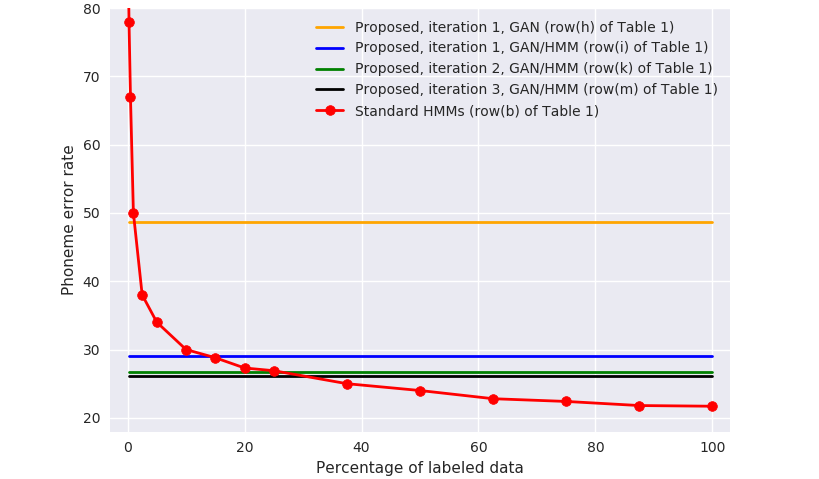}
  \caption{Comparison of the proposed approaches to standard supervised HMMs with varying quantity of labeled data.} 
  \label{fig:exp3}
\end{figure}

Here we wish to find out how the performance of the proposed approach compare to the standard supervised method when less labeled data are available. This is shown in Fig.\ref{fig:exp3}, in which the red curve is for the standard HMMs in row (b) of Table \ref{tab:performance}, whose right lower end is 21.5\% as in Table \ref{tab:performance}. Here we show how PER goes up when only a percentage of the labeled training data (4000 utterances) are available. The horizontal lines then correspond to the proposed approach GAN/HMM at iteration 3, 2, 1 and GAN alone at iteration 1, or rows (m) (k) (i) (h) of Table \ref{tab:performance}, all in the matched case. We see the proposed approaches achieved the PER of standard HMMs when roughly 30\%, 25\%, 15\% and 1\% of the labeled data are available. This also demonstrates how the harmonized GAN/HMM proposed here offered improved performance step by step.
 
\section{Conclusions}
In this work we proposed a framework to achieve unsupervised speech recognition without any labeled data. A GAN is used in which a generator and a discriminator learn form each other iteratively, and a set of HMMs is further harmonized iteratively with the GAN. Dramatically improved performance was obtained compared to the previously reported results.

\bibliographystyle{IEEEtran}
\bibliography{mybib}
\end{document}